\pdfoutput=1

\documentclass[11pt]{article}

\usepackage[final]{acl}

\usepackage{times}
\usepackage{latexsym}

\usepackage[T1]{fontenc}

\usepackage[utf8]{inputenc}

\usepackage{microtype}

\usepackage{inconsolata}

\usepackage{graphicx}

\usepackage{amsmath} 
\usepackage{amssymb} 
\usepackage{booktabs} 
\usepackage{makecell} 
\usepackage{array} 
\usepackage{tipa} 
\usepackage{multirow} 

\usepackage{fontawesome} 

\title{A Two-Step Approach for Data-Efficient French Pronunciation Learning}

\author{
 \textbf{Hoyeon Lee\textsuperscript{1}} \quad
 \textbf{Hyeeun Jang\textsuperscript{2}} \quad
 \textbf{Jong-Hwan Kim\textsuperscript{1}} \quad
 \textbf{Jae-Min Kim\textsuperscript{1}}
\\
\\
 \textsuperscript{1}NAVER Cloud \quad
 \textsuperscript{2}Université de Strasbourg
\\
 \texttt{yeon.lee@navercorp.com} \\
}

\begin{document}
\newcommand{\eunedit}[1]{\textcolor{blue}{#1}}
\newcommand{\yeonedit}[1]{\textcolor{green}{#1}}
\maketitle
\begin{abstract}
Recent studies have addressed intricate phonological phenomena in French, relying on either extensive linguistic knowledge or a significant amount of sentence-level pronunciation data.
However, creating such resources is expensive and non-trivial.
To this end, we propose a novel two-step approach that encompasses two pronunciation tasks: grapheme-to-phoneme and post-lexical processing.
We then investigate the efficacy of the proposed approach with a notably limited amount of sentence-level pronunciation data.
Our findings demonstrate that the proposed two-step approach effectively mitigates the lack of extensive labeled data, and serves as a feasible solution for addressing French phonological phenomena even under resource-constrained environments.
\end{abstract}

\section{Introduction}
Phonetic information plays a crucial role in text-to-speech systems, improving the clarity and naturalness of synthetic speech.
Grapheme-to-phoneme (G2P) relationships are typically modeled using a sizeable set of phonetic transcriptions to predict the pronunciation of out-of-vocabulary words.
However, pronunciation learning in French remains challenging due to its intricate phonetic structure and phonological phenomena such as \textit{Linking (Enchaînement)} and \textit{Liaison}.

These phenomena mediate between words by modifying phonemes and their placement~\cite{adda1999pronunciation, bybee2001frequency}.
\textit{Linking} is the articulation of a consonant-final word and its re-syllabification with the following vowel-initial word~\cite{gaskell2002perception, fougeron2003looking}.
For example, when a consonant-final word \textit{une} - [yn] precedes a vowel-initial word \textit{amie} - [a.mi], the phoneme ``n'' is resyllabified and positioned adjacent to the primary phoneme of the following word, as illustrated in the example below:
\begin{center}
    \textit{une} [yn] \textit{amie} [a.mi] $\rightarrow$ \textit{une amie} [y.\textbf{n}a.mi]
\end{center}
\textit{Liaison} refers to the pronunciation of a silent consonant-final word to its phoneme or another when it is followed by a vowel-initial word~\cite{bybee2001frequency, gaskell2002perception}.
For instance, when the determiner \textit{mes} is an initial word, its pronunciation is altered depending on the following word.
If the following word, such as \textit{frères}, begins with a consonant, the ``s'' grapheme remains silent in the corresponding pronunciation.
\begin{center}
    \textit{mes} [me] \textit{frères} [f\textinvscr \textepsilon \textinvscr] $\rightarrow$ \textit{mes frères} [me.\textbf{f}\textinvscr \textepsilon \textinvscr] \\
\end{center}
However, when it is followed by a vowel-initial word like \textit{amis}, the ``s'' is pronounced as [z].
\begin{center}
    \textit{mes} [me] \textit{amis} [ami] $\rightarrow$ \textit{mes amis} [me.\textbf{z}a.mi]
\end{center}
These phonetic modifications are influenced by various factors and contexts, leading to numerous exceptions. Such variability further adds to the complexity of addressing phonological phenomena\footnote{Further elaboration and examples of the complexities are described in Appendix~\ref{apx:complexities_of_liaison}.}.

Recent studies have focused on modeling these phonological phenomena through two main approaches: post-lexical rules (PLR)~\cite{tzoukermann1998text} and data-driven methods~\cite{pontes2010modeling, taylor2021liaison, comini2023multilingual}.
While both PLR and data-driven methods achieve decent performance, they demand deeper phonological/linguistic knowledge and a substantial amount of human-annotated sentence-level data, respectively.
Unfortunately, existing approaches without such expensive resource commitments have not been satisfactory.

\begin{figure*}[t]
    \centering
    \includegraphics[width=1.0\textwidth]{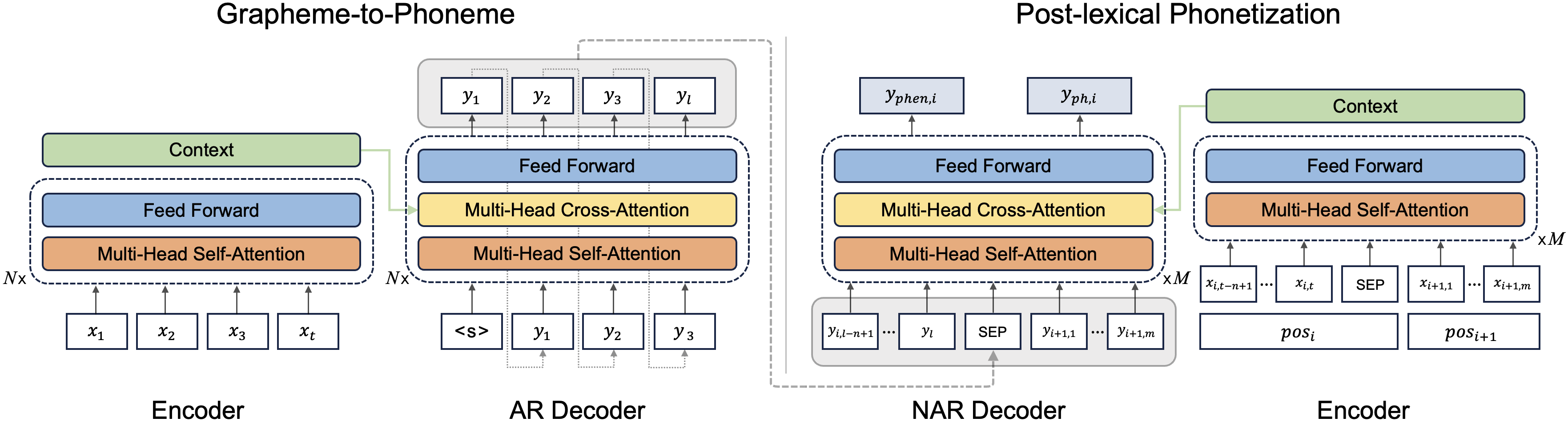}
    \caption{An overview of our proposed architecture.}
    \label{fig:architecture}
\end{figure*}

In this paper, we propose a novel two-step approach to address the challenge of learning French pronunciation with limited resources.
Specifically, we explicitly decompose the intricate and comprehensive pronunciation task into two sub-tasks: G2P conversion and post-lexical processing.
First, we leverage a large amount of easily accessible word-level pronunciation data to train the autoregressive transformer (ART)-based G2P model~\cite{comini2023multilingual, yolchuyeva2020transformer, yu2020multilingual, zhu2022byt5} and generate correct pronunciations for corresponding words.
Second, we adopt a shallow non-autoregressive transformer (NART)~\cite{gu2017non, sun2019fast} as the post-lexical phonetization model to process phonological phenomena between the pronunciations of individual words.
This model is trained on our manually constructed dataset, comprising a modest number of sentence-level examples.

Accordingly, we assess whether the proposed approach can effectively leverage a limited set of sentence-level examples to overcome the challenges, and further analyze how varying the size of these resources impacts performance.
The experimental results reveal that the proposed approach successfully addresses intricate phonological phenomena, utilizing only around 2k examples, with even 1.5k examples proving somewhat effective.

\section{Related Work}

PLR is one of the essential modules used to address phonological phenomena in the French text-to-speech front-end, yet it requires extensive phonological and linguistic knowledge to construct a comprehensive set of hand-crafted post-lexical rules.
Although only a few studies\footnote{\url{http://research.jyu.fi/phonfr/20.html}}~\cite{tzoukermann1998text} provide initial guidelines for manually constructing post-lexical rules, the implementation of their intricate interactions still necessitates the deep knowledge and substantial efforts of linguistic experts.

To alleviate this burden, data-driven approaches have also been proposed~\cite{pontes2010modeling, taylor2021liaison, comini2023multilingual}.
Unlike PLR, these approaches demonstrate performance improvements by leveraging large-scale sentence-level pronunciation datasets, even without linguistic knowledge.
~\citet{comini2023multilingual} reported using a pronunciation dataset for training, comprising about 33.7k sentences and 800k words, with the sentence-level pronunciations generated via internal front-ends.
Nevertheless, they also highlighted that, even with 33.7k sentence-level phonetic transcriptions, the absence of data specifically designed for post-lexical processing may still cause certain contexts to be missed.

\section{Proposed Method}
In this section, we describe our two-step approach to French pronunciation learning.
Our approach addresses the extensive and complex pronunciation task by explicitly decomposing it into two key sub-tasks: G2P conversion and post-lexical processing.
The overall architecture is illustrated in Figure~\ref{fig:architecture}.

\subsection{Grapheme-to-Phoneme}
To generate pronunciation for a given word, we employ a vanilla ART architecture~\cite{vaswani2017attention}.
Following the autoregressive model-based sequence-to-sequence paradigm applied in G2P~\cite{milde2017multitask, peters2017massively, yolchuyeva2020transformer, yu2020multilingual, zhu2022byt5, comini2023multilingual},
the encoder transforms the grapheme sequence $x = {\{{x}_{1}, {x}_{2}, ..., {x}_{t}\}}$ into contextual information, and the decoder generates the corresponding phoneme sequence $y = {\{{y}_{1}, {y}_{2}, ..., {y}_{l}\}}$ based on the encoder's output.
We train the ART G2P model on word-level pronunciation data, which includes \textit{$<$word, pronunciation$>$} pairs like \textit{$<$enfant, $\tilde{a}$f$\tilde{a}$$>$}.
During training, we use cross-entropy (CE) loss between the generated phonemes $\hat{y}$ and ground truth phonemes ${y}$.

\subsection{Post-lexical Phonetization}
The key intuition underlying our approach is to adopt a separate post-lexical processing module rather than directly predicting phoneme sequences covering the phonological phenomena.
The hypothesis is that learning French pronunciation, including the post-lexical phenomena, is particularly challenging when relying on a limited sentence-level dataset. This challenge is compounded by the fact that each example contains merely a few words affected by the phenomena.
This dedicated module targets post-lexical phonetization, effectively leveraging a modest number of sentence-level examples.

The post-lexical phonetization model follows a similar architecture as the G2P model of the first sub-module.
The key distinction here is using a fairly shallow, non-autoregressive architecture.
The encoder compresses the concatenated grapheme sequences of the partial grapheme sequence of the word ${x}_{i}$ and the following word ${x}_{i+1}$, along with the part-of-speech (POS) tags extracted from a pre-trained POS model.
Given that post-lexical phenomena are related to the graphemes between the immediately surrounding words, we use the final $n$ graphemes of the word ${x}_{i}$ and the initial $m$ graphemes of the following word ${x}_{i+1}$.
A [SEP] token is inserted between each partial grapheme sequence.
When the number of graphemes in each word is fewer than the pre-defined values of $n$ and $m$, a [PAD] token is added to the beginning or end of each sequence, respectively.
This approach allows the encoder to effectively capture contextual information essential for processing post-lexical phenomena, leveraging both grapheme sequences and POS tags of adjacent words.

Conditioned on the contextual information extracted from the encoder, the decoder then predicts whether a phonological phenomenon occurs ${y}_{phen}$ and the resulting phoneme ${y}_{ph}$.
We utilize word-level phoneme sequences predicted by the pre-trained G2P model as input for the decoder. 
Similar to the encoder input, the decoder's input comprises the concatenation of partial phoneme sequences, encompassing the final $n$ phonemes of the word ${x}_{i}$ and the initial $m$ phonemes of the subsequent word ${x}_{i+1}$.
[SEP] and [PAD] tokens are appended in the same manner as the encoder.

We then introduce two loss terms to train the phonetization model.
The first loss term ($\mathcal{L}_{phen}$) is the weighted binary cross-entropy (WBCE) loss used to identify the presence of the phonological phenomenon in class-imbalanced settings:
\begin{equation}
    \label{eq:plp-phen}
    \mathcal{L}_{phen} = \mathrm{WBCE}\left( \hat{y}_{phen}, {y}_{phen} \right)
\end{equation}
where ${y}_{phen}$ is the ground truth of phonological phenomenon occurrence, and $\hat{y}_{phen}$ is the predicted probability of the phonological phenomenon.
The second loss term ($\mathcal{L}_{ph}$) is the CE loss, used for predicting pronunciation variation as follows:
\begin{equation}
    \label{eq:plp-ph}
    \mathcal{L}_{ph} = y_{phen} \cdot \mathrm{CE}\left( \hat{y}_{ph}, {y}_{ph} \right)
\end{equation}
where ${y}_{ph}$ is the ground truth of the phoneme at locations where phonological phenomenon may occur.
The key characteristic is adding a simple selector that is selectively activated based on the occurrence of a phenomenon, as follows:
\begin{equation}
    \label{eq:plp-phen-case}
    y_{phen, i} = 
        \begin{cases} 
        1, & \text{if post-lexical phenomenon} \\
        0, & \text{otherwise}
        \end{cases}
\end{equation}
Overall, the loss of the phonetization model is the sum of $\mathcal{L}_{ph}$ and $\mathcal{L}_{phen}$, as follows:
\begin{equation}
    \label{eq:plp}
    \mathcal{L}_{plp} = \mathcal{L}_{ph} + \mathcal{L}_{phen}
\end{equation}

\section{Experiment}

\begin{table}[t]
  \centering
  {\footnotesize
  \begin{tabular}{lrr}
    \toprule
    & \multicolumn{1}{c}{\textit{N}} & \multicolumn{1}{c}{\textit{SD}} \\
    \midrule
    Examples & 2,645 & - \\
    Examples w/ phonological phenomena & 2,107 & - \\
    Examples w/o phonological phenomena\phantom{4} & 538 & -\\
    Avg. words & 12.27 & 2.96 \\
    Avg. \textit{Liaison} & 0.60 & 0.76 \\
    Avg. \textit{Linking} & 0.83 & 0.90 \\
    \bottomrule
  \end{tabular}
  }
  \caption{Statistics of sentence-level pronunciation dataset. The number of examples and the average number of words and occurrences of phonological phenomena per example.}
  \label{tab:dataset}
\end{table}

\subsection{Datasets}
First, we collect pronunciation data at both the word and sentence levels.
Following \citet{comini2023multilingual}, we gather word-level pronunciation data using an internal pronunciation dictionary, which contains 106,857 unique entries, each with phonetic transcriptions denoted using X-SAMPA notation.
Furthermore, we collect 2,645 sentence-level data from various domains, including news, social media posts, and the Multilingual LibriSpeech (MLS)~\cite{Pratap2020MLSAL} dataset.
Each example is manually annotated by a French linguistic expert using the same X-SAMPA phonetic transcription as the word-level data, including phonological phenomena like \textit{Liaison} and \textit{Linking}\footnote{The details of the phonetic transcription and example are provided in Appendix~\ref{apx:transcription_example}}.
Each sentence consists of an average of 12.27 (\textit{SD} = 2.96) words, and the average frequency of phonological phenomena occurrences per sentence is 1.43 (\textit{SD} = 1.13).
Table~\ref{tab:dataset} reports the detailed statistics of the sentence-level dataset.

In addition to resource creation, we apply only essential preprocessing steps and curation criteria to avoid examples that may negatively impact the performance of our proposed method, while preserving the original context and natural structure of the data.
The overall preprocessing techniques and curation criteria applied to the dataset are described in Appendix~\ref{apx:data_preprocessing}.

\subsection{Experimental Setup}
We split the word-level dataset into training (85\%), validation (5\%), and test (10\%) sets for the G2P model training.
For the post-lexical phonetization model, the number of training instances $k$ varies from 2,045 to 512, decreasing in 25\% intervals, with 300 examples for both the validation and test sets.
To evaluate the G2P and post-lexical phonetization models, we use the following metrics: phoneme error rate (PER), word error rate (WER), and {Acc}$_{plp}$ (Appendix ~\ref{apx:metrics}).

Within our proposed method, the ART G2P model is mainly implemented following the setup described in~\cite{zhu2022byt5}.
The transformer encoder and decoder consist of 8 layers each, with 8 self-attention heads, 512-dimensional embeddings, and 2048 feed-forward dimensions, resulting in 58.9M parameters.
For the post-lexical phonetization model, we employ a shallow NART architecture. This architecture consists of just 2 transformer layers, each with 8 self-attention heads, and 512-dimensional embeddings, resulting in 14.8M parameters only.
The total number of parameters for the proposed architecture is around 73.7M.
Additional details of other configurations are provided in Appendix~\ref{apx:configs}.

Building on this foundation, we compare our proposed method to the ART-based G2P model, which serves as our baseline due to its renowned superior performance in existing approaches.
This model is trained in three distinct settings: using word-level data~\cite{yolchuyeva2020transformer}, sentence-level data, and a combination of both~\cite{comini2023multilingual}.
To ensure a fair comparison with the proposed method, we employ a 10-layer ART G2P model and follow the same procedure to identify optimal hyperparameters in all settings.

\subsection{Results and Analysis}

\begin{table}
    \centering
    \resizebox{\linewidth}{!}{%
    \begin{tabular}{clccc}
        \toprule
        Annotation & Phonological Case & {Acc}$_{plp}$ $\uparrow$ & PER $\downarrow$ & WER $\downarrow$ \\
        \midrule
        \multirow{4}{*}{Word}    & Whole sentence & \textbf{84.92} & \textbf{10.64} & \textbf{24.68} \\ 
        & Phonological phen.    & 0.00 & \textbf{12.40} & \textbf{29.03} \\
        & \textit{Liaison}      & 0.00 & \textbf{11.87} & \textbf{28.66} \\
        & \textit{Linking}      & 0.00 & \textbf{14.58} & \textbf{33.65} \\
        \midrule
        \multirow{4}{*}{Sentence}   & Whole sentence & 14.37 & 136.48 & 95.04 \\
        & Phonological phen.        & 13.23	& 135.71 & 95.07 \\
        & \textit{Liaison}          & 12.68 & 133.60 & 95.45 \\
        & \textit{Linking}          & 13.69 & 136.10 & 95.37 \\
        \midrule
        \multirow{4}{*}{Word/Sentence}   & Whole sentence & 77.85 & 29.91 & 34.39 \\
        & Phonological phen.        & \textbf{55.38} & 32.79 & 37.21 \\
        & \textit{Liaison}          & \textbf{67.32} & 34.00 & 37.24 \\
        & \textit{Linking}          & \textbf{45.23} & 33.10 & 38.72 \\
        \bottomrule
    \end{tabular}
    }
    \caption{Evaluation of the baseline models with different types of pronunciation datasets.}
    \label{tab:baseline}
\end{table}

\begin{table*}[]
    \centering
    \resizebox{\linewidth}{!}{%
    \begin{tabular}{lcccccccccc}
        \toprule
        & \multicolumn{2}{c}{25\% of \textit{Full}} & \multicolumn{2}{c}{50\% of \textit{Full}} & \multicolumn{2}{c}{75\% of \textit{Full}} & \multicolumn{2}{c}{\textit{Full}} \\ \cmidrule(lr){2-3} \cmidrule(lr){4-5} \cmidrule(lr){6-7} \cmidrule(lr){8-9}
        & {Acc}$_{plp}$ $\uparrow$ & PER/WER $\downarrow$ & {Acc}$_{plp}$ $\uparrow$ & PER/WER $\downarrow$ & {Acc}$_{plp}$ $\uparrow$ & PER/WER $\downarrow$ & {Acc}$_{plp}$ $\uparrow$ & PER/WER $\downarrow$ \\
        \midrule
        Whole sentence   & 93.80\scriptsize{(+16.0)} & 5.67/14.40 & 93.93\scriptsize{(+16.1)} & 5.69/13.54 & 94.52\scriptsize{(+16.7)} & 5.27/12.80 & \textbf{95.52\scriptsize{(+17.7)}} & \textbf{4.79}/\textbf{11.47} \\
        Phonological phen.   & 69.51\scriptsize{(+14.1)} & 6.20/15.57 & 71.30\scriptsize{(+15.9)} & 6.24/14.55 & 80.99\scriptsize{(+25.6)} & 5.64/13.46 & \textbf{83.86\scriptsize{(+28.5)}} & \textbf{5.18}/\textbf{12.11} \\
        \phantom{12}\textit{Liaison}   & 63.73\scriptsize{(-3.6)} & 6.06/15.49 & 67.16\scriptsize{(-0.2)} & 6.20/15.17 & 68.16\scriptsize{(+0.8)} & \textbf{5.42}/13.71 & \textbf{77.56\scriptsize{(+10.2)}} & 5.53/\textbf{13.06} \\
        \phantom{12}\textit{Linking}   & 74.38\scriptsize{(+29.2)} & 6.64/16.92 & 74.79\scriptsize{(+29.6)} & 6.52/15.42 & 91.74\scriptsize{(+46.5)} & 5.73/13.81 & \textbf{89.21\scriptsize{(+44.0)}} & \textbf{5.11}/\textbf{12.11} \\
        \bottomrule
    \end{tabular}
    }
    \caption{Evaluation of the proposed approach with varying numbers of sentence-level instances. \textit{Full} denotes using all sentence-level instances. Differences from the baseline, trained on word/sentence-level data, are in parentheses.}
    \label{tab:plp-eval}
\end{table*}

We evaluate the proposed approach alongside several baseline models, by breaking down the analysis into four distinct cases: Whole sentence, Phonological phenomena, \textit{Liaison}, and \textit{Linking}.
\paragraph{Can the baseline models address phonological phenomena?}
We train the naive ART G2P model on three distinct sets of pronunciation data, with the experimental results detailed in Table~\ref{tab:baseline}.
As expected, the model trained solely on word-level phonetic transcription data was completely unable to address any phonological phenomena, resulting in an ${\text{Acc}}_{plp}$ of 0\% in all phonological phenomena cases.
Despite this, when leveraging around 100k entries of large-scale word-level data for training, the model showed adequate performance in PER and WER, aligning with the findings of previous studies. 
For the model trained on sentence-level pronunciation data, we can observe a slight improvement in addressing phonological phenomena with an ${\text{Acc}}_{plp}$ of 13.23\%.
However, due to the limited number of training examples, the overall performance substantially declined.
In contrast, compared to using sentence-level data alone, combining a large amount of word-level data with a small amount of sentence-level data led to a considerable improvement.
PER and WER significantly decreased in all cases, reaching 29.91\% and 34.39\%, respectively.
${\text{Acc}}_{plp}$ also improved relative to other baselines, reaching 55.38\% in the Phonological phenomena case; however, it remains markedly inferior compared to the Whole sentence.

\paragraph{How effective is the two-step approach in addressing phonological phenomena?}
Initially, we evaluate the G2P model of the proposed approach to generate correct word-level pronunciations that serve as inputs for the subsequent post-lexical phonetization model.
Interestingly, the G2P model demonstrates performance on par with the baseline model, achieving a PER of 9.78\%, a WER of 24.95\%, and an ${\text{Acc}}_{plp}$ of 84.95\% in the Whole sentence case.
Based on the aforementioned G2P model, we evaluate our proposed approach, with the overall results illustrated in Table~\ref{tab:plp-eval}.
As anticipated, by utilizing the entire dataset, the \textit{Full} model achieved the best performance in all cases, with the only exception of the PER in the \textit{Liaison} case, which used 75\% of the \textit{Full} dataset.
Relative to the baseline, trained on sentence-level data capable of handling phonological phenomena to a certain extent, we can observe a considerable improvement in all metrics.
Most notably, the ${\text{Acc}}_{plp}$ for phonological phenomena cases showed a remarkable increase to 83.86\%, up by an average of 28.5\%, representing a substantial rise from Whole sentence.
This result demonstrates the notable effectiveness of the two-step approach in addressing phonological phenomena, even with limited data.

\paragraph{How many sentence-level phonetic transcriptions are needed for our approach to work well?}
Reflecting the application in certain scenarios, possibly constrained by the more extreme scarcity of consistently labeled sentence-level data~\cite{lee2023cross}, we conduct a further empirical analysis exploring how resource size affects performance across all cases.
As illustrated in Table~\ref{tab:plp-eval}, decreasing the number of training instances resulted in a progressive decline in performance across all cases, relative to the \textit{Full} model.
Although using less than half of the entire dataset led to improvements over the baseline in most cases, a marginal performance drop from 0.2\% to 3.6\% was observed in the \textit{Liaison} case.
This implies that relying on an excessively small number of examples may miss certain contexts, proving insufficient to capture the entire spectrum of phonological variation on contextual factors~\cite{de2003liaisons, kondo2012liaison, encreve1983liaison}.
In contrast, it is noteworthy that utilizing more than 75\% of the entire dataset leads to a noticeable improvement, surpassing 80\% in the Phonological phenomena case. Specifically, the \textit{Linking} case achieves around 90\% in ${\text{Acc}}_{plp}$.
Drawing from these results, we conjecture that addressing phonological phenomena requires a minimum of about 1.5k sentence-level examples.

\section{Conclusion}
In this paper, we present an effective two-step approach for French pronunciation learning.
Our approach alleviates the burden of extensive resources by decomposing the intricate and comprehensive pronunciation task into two sub-tasks, thereby facilitating greater leverage of a modest number of sentence-level examples.
The empirical analysis demonstrates the efficacy of our proposed approach in addressing phonological phenomena even in resource-constrained environments.


\section{Limitations}
This study has a few important limitations.
We employed closed-source datasets.
Although some open-source word-level pronunciation datasets exist, such as the one provided by \citet{zhu2022byt5}, which contains approximately 250k French word-pronunciation pairs and is significantly larger than ours, we chose not to use it due to the presence of noise.
Moreover, finding a publicly available dataset designed for post-lexical processing has been significantly challenging, as even sentence-level phonetic transcriptions reflecting phonological phenomena are not available.

Our manually constructed sentence-level dataset contains a smaller amount of annotated data compared to previous research. This limited size may not completely capture the generalizability of the entire spectrum of French phonological phenomena described in Appendix~\ref{apx:complexities_of_liaison}.
While this smaller dataset may result in some missing contexts, it represents our main contribution. Despite being a data-driven method, we achieved significant results using only less than 2k sentence-level pronunciation data.

The post-lexical phonetization model in our proposed method predicts a single resulting phoneme, which serves as the ground truth at locations where a phonological phenomenon may occur, though it cannot address the few cases involving more than one phonetic shift (including vowel changes) beyond the final grapheme of the first word in a sequence.
For instance, in certain cases like \textit{bon ami}, a phonetic shift occurs not at the final grapheme, but at the second-to-last grapheme:
\begin{center}
    \textit{bon} [b\textipa{\~O}] \textit{ami} [ami] $\rightarrow$ \textit{bon ami} [b\textbf{o}.na.mi]
\end{center}
This represents a vowel shift, where the nasal vowel [\textipa{\~O}] transitions to the oral vowel [o] during the \textit{liaison} process.

We evaluated the proposed approach by comparing it with data-driven approaches, rather than a PLR method. The major challenge in comparing with a PLR method is that it requires significant expertise, as well as substantial temporal and financial costs for the manual construction of numerous post-lexical rules and their interactions.
Therefore, we focused on comparisons with data-driven approaches, which may be more feasible solutions in resource-constrained environments.


\bibliography{custom}


\appendix
\section{Complexities of French Liaison}
\label{apx:complexities_of_liaison}
Historically rooted in the pronunciation of final consonants, French \textit{liaison} has evolved into a complex linguistic feature influenced by socio-cultural factors~\cite{adda1999pronunciation}.
Consequently, it is categorized as obligatory, optional, or forbidden \textit{liaison} based on the morpho-syntactic context~\cite{durand2008french}.
For instance, while it is obligatory for the initial word \textit{quand} with any subsequent vowel-initial word, the forbidden configuration in which \textit{et} serves as the initial word indicates that regardless of the subsequent vowel-initial word, \textit{liaison} will not occur.
Additionally, the context of optional \textit{liaison} may vary on subjective choices made by speakers, which depend on stylistic, socio-linguistic, and situational factors~\cite{durand2008french, de2003liaisons}.

Exceptions arising from fixed expressions further compound the complexity of \textit{liaison}~\cite{laks2005liaison, bybee2005liaison}.
For instance, in the fixed expression \textit{accent aigu}, the final grapheme ``t'' of \textit{accent} forms a \textit{liaison} with the initial vowel of \textit{aigu}:
\begin{center}
    \textit{accent} [ak.s\~a] \textit{aigu} [e.gy] $\rightarrow$ \\ \textit{accent aigu}  [ak.s\~a.\textbf{t}e.gy]
\end{center}
Despite the general rule that typically prevents \textit{liaison}, such exceptions enforce the phonetic link, thus increasing the complexity of addressing \textit{liaison}.

\section{Experimental Details}
\subsection{Phonetic Transcription Example}
\label{apx:transcription_example}
The following is an example of a sentence-level phonetic transcription we collected. This example includes word-level phoneme sequences reflecting the phonological phenomena in the sentence.
\begin{itemize}
    \item Sentence: \textit{Un enfant innocent a oublié sa petite envelope.}
    \item Phonetic transcription: 9\texttt{\textasciitilde} \^{} nA\texttt{\textasciitilde}fA\texttt{\textasciitilde} / inOsA\texttt{\textasciitilde} / a / ublije / sa / p@ti \^{} tA\texttt{\textasciitilde}vlOp
\end{itemize}
If no phonological phenomenon occurs between words, a slash (/) is inserted between the word-level phoneme sequences. However, if a phonological phenomenon is present, caret (\^{}) is added.

\subsection{Data Preprocessing}
\label{apx:data_preprocessing}
To ensure our data retain their original meaning and naturalness, we apply only essential preprocessing steps: (i) character case folding, (ii) removing of special characters (e.g., HTML tags, links, emojis), and (iii) replacing punctuation marks with `\#'.
Crucially, apostrophes (') and hyphens (-) between words are preserved to avoid distorting the original meaning.
Furthermore, our pronunciation datasets are composed of entries that meet the following criteria: (i) a complete sentence structure, (ii) a minimum of four words per sentence, (iii) words represented by no more than 32 graphemes or phonemes, and (iv) fewer than 192 total characters in the sentence.

\subsection{Metrics}
\label{apx:metrics}
PER is the Levenshtein distance between the predicted and reference phoneme sequences, divided by the reference's length, and WER is the percentage of words with predicted phoneme sequences mismatched with the reference.
For a focused evaluation of the post-lexical phonetization model's ability to address phonological phenomena, the ${\text{Acc}}_{plp}$ metric is employed, representing the accuracy of phonemes at locations where phonological phenomena may occur.

\subsection{Configuration Details}
\label{apx:configs}

\paragraph{Hyperparameters}
Following \citet{lee2023cross}, we conducted a grid search across several hyperparameters, exploring a variety of combinations to ensure optimal performance.
Thus, based on the validation set performance, we selected the following settings: the AdamW optimizer with a learning rate of 1e-4, a dropout rate of 0.1, 100 epochs, and early stopping after 5 epochs.

\paragraph{Post-lexical Phonetization}
We set \textit{n}=5 and \textit{m}=5 for constructing input sequences of encoder/decoder, as preliminary experiments showed these values to be optimal for effective training and computational efficiency.

\paragraph{Part-of-Speech Tagging}
Drawing on insights from the previous studies~\cite{de2003liaisons, taylor2021liaison}, we adopt the open-source pre-trained French Part-of-Speech (POS) tagging model\footnote{\url{https://huggingface.co/gilf/french-camembert-postag-model}\label{guide}}.
The model is fine-tuned on the Free French Treebank dataset\footnote{\url{https://github.com/nicolashernandez/free-french-treebank}} consisting of 29 POS tags~\cite{crabbe2008experiences}.
The French POS model extracts POS tags from the input text, which are then utilized as an auxiliary input for the post-lexical phonetization model. Its parameters are frozen during the training.

\end{document}